\documentclass[9pt,journal,final]{IEEEtran}
\makeatletter
\renewcommand\normalsize{%
\pdfoutput=1
\@setfontsize\normalsize\@xpt\@xiipt
\abovedisplayskip 4\p@ \@plus2\p@ \@minus5\p@
\abovedisplayshortskip \z@ \@plus3\p@
\belowdisplayshortskip 6\p@ \@plus3\p@ \@minus3\p@
\belowdisplayskip \abovedisplayskip
\let\@listi\@listI}
\makeatother

\usepackage{float}
\usepackage{color}
\usepackage{graphicx}
\graphicspath{{./Figures/}}
\usepackage{amsmath}
\usepackage[justification=centering]{caption}

\usepackage{verbatim}

\usepackage{multirow}
\usepackage{makecell}

\usepackage{amsmath}

\usepackage{amssymb}
\usepackage{algorithmic}
\usepackage{algorithm}
\usepackage{booktabs}
\usepackage{cite}
\usepackage[colorlinks,linkcolor=blue,citecolor=blue,urlcolor=black]{hyperref}

\begin{document}
\title{Wind Power Forecasting Considering Data Privacy Protection: A Federated Deep Reinforcement Learning Approach}

\author{Yang~Li,~\IEEEmembership{Senior Member,~IEEE,}
       Ruinong~Wang, Yuanzheng Li,~\IEEEmembership{Member,~IEEE}, Meng~Zhang, Chao Long 
       
\thanks{Y. Li is with the Key Laboratory of Modern Power System Simulation and Control and Renewable Energy Technology, Ministry of Education, Northeast Electric Power University, Jilin 132012, China (e-mail: liyang@neepu.edu.cn).
\par R. Wang is with the State Grid Jilin Electric Power Co., Ltd. Jilin Power Supply Company, Chuanying District, Jilin 132001, China (e-mail: 624895223@qq.com)
\par Y. Z. Li is with the School of Artificial Intelligence and Automation, Huazhong University of Science and Technology, Wuhan 430074, China (e-mail: Yuanzheng\_Li@hust.edu.cn)
\par M. Zhang is with the School of National Key Laboratory of Science and Technology on Vessel Integrated Power System, Naval University of Engineering, Wuhan 430033, China (e-mail: 964285846@qq.com)
\par C. Long is with the School of Water, Energy and Environment, Cranfield University, UK (e-mail: Chao.Long@cranfield.ac.uk).
}
}

\maketitle
\begin{abstract}
    In a modern power system with an increasing proportion of renewable energy, wind power prediction is crucial to the arrangement of power grid dispatching plans due to the volatility of wind power. However, traditional centralized forecasting methods raise concerns regarding data privacy-preserving and data islands problem. To handle the data privacy and openness, we propose a forecasting scheme that combines federated learning and deep reinforcement learning (DRL) for ultra-short-term wind power forecasting, called federated deep reinforcement learning (FedDRL). Firstly, this paper uses the deep deterministic policy gradient (DDPG) algorithm as the basic forecasting model to improve prediction accuracy. Secondly, we integrate the DDPG forecasting model into the framework of federated learning. The designed FedDRL can obtain an accurate prediction model in a decentralized way by sharing model parameters instead of sharing private data which can avoid sensitive privacy issues. The simulation results show that the proposed FedDRL outperforms the traditional prediction methods in terms of forecasting accuracy. More importantly, while ensuring the forecasting performance, FedDRL can effectively protect the data privacy and relieve the communication pressure compared with the traditional centralized forecasting method. In addition, a simulation with different federated learning parameters is conducted to confirm the robustness of the proposed scheme.

\end{abstract}
    
\begin{IEEEkeywords}
    Wind power forecasting, Data openness and sharing, Privacy protection, Deep reinforcement learning, Federated learning, Uncertainty modeling.      
\end{IEEEkeywords}

\section{Introduction}\label{sec.intro}
\IEEEPARstart{U}{nder} the dual pressure of the gradual exhaustion of non-renewable energy and the increasingly prominent ecological and environmental problems, how to achieve a clean and sustainable energy supply has become a critical issue that needs to be addressed \cite{li2018two}. As a promising solution, wind power has been penetrating the power system in recent years. Accurate forecasting methods can mitigate the impact of the randomness and volatility of wind power generation on the operation and dispatch of the power system, and accelerate the large-scale grid connection of wind farms\cite{arora2022probabilistic}\cite{buhan2015multistage}. Protecting the data security and privacy is becoming one of the most important challenges to enable efficient data openness and sharing among stakeholders in modern energy systems. With the continuous improvement of privacy protection awareness, it is also crucial to find solutions to protect users' data privacy while ensuring the performance of the forecasting model \cite{zeadally2013towards}.
\subsection{Literature Review}
Wind power forecasting methods can be summarized as physical and statistical prediction methods according to their modeling approaches. The main idea of the physical prediction method is to convert the data output by the numerical weather prediction (NWP) system to the wind turbine data, and then make the wind power prediction based on the predicted wind speed and the power curve of the wind turbine\cite{sweeney2020future}. This method has relatively a low prediction accuracy and is generally applied to newly established wind farms that lack of massive historical data. Statistical forecasting methods are mainly based on a large amount of historical data and using different statistical models for wind power forecasting, including time series analysis and machine learning algorithms. 1) Time series analysis methods: Ref. \cite{rajagopalan2009wind} uses autoregressive and moving average (ARMA) models for wind power prediction; to improve the prediction accuracy, \cite{liu2021short} proposes a new seasonal autoregressive integrated moving average (ARIMA) for short-term wind speed forecasting; Ref. \cite{cassola2012wind} uses the famous kalman filtering (KF) to predict wind speed and wind power. However, traditional time series prediction methods have very rigorous requirements on parameters model, and the choice of parameters model can directly determine the model performance. 2) Machine learning algorithms: In \cite{hu2015short}, a new framework called heteroscedastic support vector regression is designed for wind speed prediction, and a brand new method for wind speed prediction combines improved empirical mode decomposition (EMD) and generic algorithm-back propagation (GA-BP) neural network are proposed in \cite{wang2016wind}. With the rise of deep learning, more sophisticated neural networks are applied to predict wind speed and wind power. Ref. \cite{khodayar2017rough} designs a deep neural network (DNN) architecture with stacked autoencoder and stacked denoising autoencoder for wind speed forecasting; \cite{wang2018deep} adopts k-means clustering algorithm to process NWP data and then uses deep belief network for short-term wind power prediction. The traditional supervised learning prediction model has a good prediction performance but requires high-quality data for training. In recent years, deep reinforcement learning, which combines perception and decision-making capabilities, has been applied for prediction problems to achieve a good performance. In  \cite{zhang2021novel}, a novel forecasting method based on deep deterministic policy gradient (DDPG) is designed for load forecasting, and a hybrid ensemble deep reinforcement learning model is proposed in \cite{liu2020new} for short-term wind speed forecasting. The basis for machine learning methods to obtain accurate prediction results is the massive amount of training data, which is very difficult for some newly built wind farms. At present, major methods generally use centralized training technology, which requires a central organization to collect data. However, due to some commercial factors, wind farms belonging to different stakeholders are not willing to share private data with others \cite{li2018optimal567}, so the central organization has to consider the issue of privacy protection when collecting and storing data.

\begin{table}[t]
  \centering
    \begin{tabular}{|ll|}
    \hline
          &  \\          
    \textbf{Nomenclature} &  \\
          &  \\
    \textbf{Acronyms} &  \\
    DRL   & Deep Reinforcement Learning \\
    FedDRL & Federated Deep Reinforcement Learning \\
    DDPG  & Deep Deterministic Policy Gradient  \\
    NWP   & Numerical Weather Prediction \\
    ARMA  & Autoregressive and Moving Average \\
    ARIMA & Autoregressive Integrated Moving Average \\
    KF    & Kalman Filtering \\
    EMD   & Empirical Mode Decomposition  \\
    GA-BP & Generic Algorithm-Back Propagation \\
    DNN  & Deep Neural Network  \\
    ADMM  & Alternate Direction Method of Multipliers \\
    MLR   & Mixed Logistic Regression \\
    BPNN  & Back Propagation Neural Network \\
    RDPG  & Recurrent Deterministic Policy Gradient \\
    NMAE  & Normalized Mean Absolute Error \\
    NRMSE & Normalized Root Mean Square Error \\
    NREL  & National Renewable Energy Laboratory \\
          &  \\        
    \textbf{Symbols} &  \\
    \textit{s} & State \\
    \textit{a} & Action \\
    \multicolumn{1}{|p{6.875em}}{\textit{r}} & Reward \\
    $\pi$ & Policy \\
    \multicolumn{1}{|p{6.875em}}{$\gamma$} & Discount factor \\
    $\theta ^{\mu}$ & Parameters of Main Actor network \\
    $\theta ^ Q$ & Parameters of Main Critic network \\
    $\theta ^{\mu '}$ & Parameters of Target Actor network \\
    $\theta ^ {Q'}$ & Parameters of Target Critic network \\
    $\varsigma$ & Soft update factor \\
    $N$ & Total number of clients \\
    $n$ & client number\\
    $\psi_C$ & Performance metrics of the centralized model \\
    $\psi_F$ & Performance metrics of the federated model \\
    $T$ & Size of minibatch\\
    $P^w$  & Wind power \\
    \textit{j} & Number of selected time periods \\
    \textit{V} & Total number of samples \\
    \textit{y} & Actual value \\
    \textit{p} & Prediction result \\
    $y^n$  & Normalized actual value \\
    $p^n$ & Normalized prediction result \\
    \textit{M} & Local episode  \\
    \textit{W} & Global epoch \\
    \textit{K} & Synchronization interval \\
    \textit{E} & Client ratio \\
    \textit{S} & Central server \\
    $L_C$ & Network load of the centralized forecasting method \\
    $L_F$ & Network load of FedDRL \\
    \textit{D} & Size of the private data \\
    \textit{I} & Size of the model information \\
    \textit{h} & Hops between each client and the central server \\
    \textit{U } & Networking load gain \\
    $p$ & Number of lag observations \\
    $d$ & Degree of differencing \\
    $q$ & Size of the moving average window\\
    \hline
    \end{tabular}%
  \label{tab:addlabel}%
\end{table}%

In recent years, many countries have promulgated relevant laws and regulations on data privacy protection to supervise the storage and application of data, and some researches using distributed structures have discussed the issue of data privacy protection \cite{voigt2017eu}. In \cite{zhang2018distributed}, a privacy protection method is proposed for wind power probabilistic forecasting which adopts the alternate direction method of multipliers (ADMM) structure. However, the central node in this method can recover the private data which may cause a confidentiality breach, resulting in the leakage of data privacy. To reduce the occurrence of this kind of privacy leakage issue, some research add random noise to the original data or coefficients when using the ADMM structure for renewable energy prediction \cite{dwork2010differential}, \cite{zhang2016dynamic}. but this method may cause a decrease in the prediction performance \cite{gonccalves2021critical}. With the continuous increase of wind farms around the world, accurate wind power prediction is more important for the stable operation and scheduling of the power system \cite{li2022deep} \cite{li2021optimal678}, which requires a large amount of historical data as support, and these data contain a lot of user privacy. In this context, it makes sense to explore a method that can both guarantee prediction accuracy and data privacy protection.

Existing research contributes magnificently to the wind power forecasting problem, however, there are still the following research gaps to be solved:
\begin{enumerate}
    \item Prediction accuracy is extremely important to the problem of wind power prediction. Discovering a method to upgrade the performance of the forecasting model is necessary. Most of the current research uses centralized forecasting methods, which may lead to some data security and privacy protection issues.
    \item Massive high-quality historical data is an essential condition for obtaining a forecasting model with good performance. However, due to some commercial factors, wind farms with different stakeholders are reluctant to share their private data. How to solve the data island problem, that is, to break the phenomenon that the data of different wind farms are stored independently and isolated from each other, which is not considered in most existing research.
    \item For wind power forecasting, the centralized prediction methods require a central organization with powerful computing ability and large storage capacity, and the data is prone to leakage when it is collected and stored.
\end{enumerate}

\subsection{Contribution of This Paper} 
To fill the above-mentioned gaps, a federated deep reinforcement learning (FedDRL) for ultra-short-term wind power forecasting is proposed in this paper. The main contributions of this paper are the following threefold:
\begin{enumerate}
    \item This paper uses the deep deterministic policy gradient (DDPG) algorithm in deep reinforcement learning (DRL) as the basic prediction model for ultra-short-term wind power forecasting which can improve the forecasting accuracy compared with the traditional forecasting methods. In addition, we combine automatic machine learning with DRL for hyperparameters selection to simplify the deployment of the forecasting model.
    \item To handle the data privacy and openness, we propose a FedDRL forecasting scheme that combines federated learning and DRL for ultra-short-term wind power forecasting of multiple wind farms. To the best of our knowledge, this paper is the first study to use federated learning for wind power forecasting. The proposed forecasting scheme can not only solve the problem of data islands but also has the advantage of privacy-preserving and reducing communication pressure under the premise of ensuring high-quality forecasting performance compared with the most used centralized forecasting methods.
    \item We use real-world historical data for simulations to verify the effectiveness of the proposed FedDRL scheme and prove that the scheme has superior forecasting performance than traditional prediction methods and the communication pressure is significantly mitigated. In addition, we also design experiments with different parameter settings to confirm the robustness of the proposed scheme.
\end{enumerate}

\subsection{Organization of This Paper} The organization of the paper is as follows: Section \ref{sec.DDPG&FED} contains the basic principles of DDPG and federated learning. Section \ref{sec.PROPOSED METHODOLOGY} introduces the methodology of the proposed FedDRL, including the transformation of the prediction problem, the structure of FedDRL, and the steps of using the proposed FedDRL for wind power prediction. Section \ref{sec.CASE STUDY} will show the results of case studies in detail and give specific conclusions and future works in the Section \ref{sec.conclusion}.

\section{RELATED PRELIMINARY WORKS}\label{sec.DDPG&FED}
This section introduces the basic theoretical knowledge related to the proposed FedDRL scheme, including a detailed description of the DDPG algorithm and the federated learning framework.
\subsection{Deep Deterministic Policy Gradient}
DRL is a branch of machine learning, which combines deep learning and reinforcement learning. As an algorithm of deep reinforcement learning, DDPG has an efficient and stable ability in solving continuous action space problems \cite{mnih2015human}.

DDPG is an algorithm based on the Actor-Critic framework which has four neural networks. Both the Actor and Critic contain a main network and a target network. The role of the target network is to stabilize the training of the algorithm.

The Actor part makes the decision of action $a$ based on the current state $s$, and after interacting with the environment, it obtains the feedback of the next state $s'$ and the reward $r$. The purpose of the Actor part is to obtain the policy with the highest total reward under a certain goal by continuously communicating with the environment, which is expressed as
\begin{equation}\label{eq.object function}
    \begin{aligned}
       J({\theta ^\mu }) = {E_{{\theta ^\mu }}}[{r_1} + \gamma {r_2} + {\gamma ^2}{r_3} + ... + {\gamma ^{i - 1}}{r_i}]
    \end{aligned}
\end{equation}
where $\gamma$ is the discount factor, $\theta^\mu$ is the parameters of the main actor network $a=\pi(s|\theta^\mu)$.

The main actor network is updated by the stochastic gradient method, and the gradient information can be approximated as follows during the update process \cite{silver2014deterministic}.
\begin{equation}\label{eq.update main actor network}
    \begin{aligned}
       \nabla _{{\theta ^\mu }}J \approx \frac{1}{T}\sum\limits_i {[\nabla {}_aQ(s,a|{\theta ^Q}){|_{s = {s_i},a = \pi ({s_i})}}\nabla {}_{{\theta ^\mu }}\pi (s|{\theta ^\mu })|_{s = {s_i}}]}
    \end{aligned}
\end{equation}
where $T$ is the size of the minibatch, $\theta^Q$ is the parameters of the main critic network $Q(s,a|\theta^Q)$.

After the Actor part makes a decision, the Critic part evaluates it and updates the main critic network with the goal of minimizing the loss function, the loss function is
\begin{equation}\label{eq.loss c}
    L({\theta ^Q}) = {E_{s,a,r,s' \sim D}}\left[ {{{\left( {TargetQ - Q(s,a|{\theta ^Q})} \right)}^2}} \right]
\end{equation}
\begin{equation}
    TargetQ = r + \gamma Q'(s',\pi (s'|{\theta ^{\mu '}})|{\theta ^{Q'}})
\end{equation}
where $Q'$ denotes the target Q-value function; $\theta ^{\mu '}$ 
is the target actor network parameters and $\theta ^{Q'}$ is the target critic network parameters. The gradient information is updated by
\begin{small}
\begin{equation}\label{eq.update the main critic network}
    \begin{aligned}
     \nabla _{{\theta ^Q}}L = {E_{s,a,r,s'\sim D}}\left[ {\left( {TargetQ - Q(s,a|{\theta ^Q})} \right){\nabla _{{\theta ^Q}}}Q(s,a|{\theta ^Q})} \right]
    \end{aligned}
\end{equation}
\end{small}

For the target networks, DDPG uses a soft method to update its parameters.
\begin{equation}\label{eq.update target networks}
    \begin{aligned}
     \begin{array}{l}
{\theta ^{Q'}} \leftarrow \varsigma {\theta ^Q} + (1 - \varsigma ){\theta ^{Q'}}\\
{\theta ^{\mu '}} \leftarrow \varsigma {\theta ^\mu } + (1 - \varsigma ){\theta ^{\mu '}}
\end{array}
    \end{aligned}
\end{equation}
where $\varsigma$ is the soft update factor.

To overcome the problems of correlated data and non-stationary distribution of empirical data, DDPG introduces Experience Replay to store the information obtained by each Actor in a buffer, which improves the utilization of data and enhances the training effect.

\subsection{Federated Learning}

Federated learning, as a machine learning framework, is first proposed by the Google R$\&$D team and applied to the keyboard prediction model for mobile phones. Compared with traditional centralized training methods\cite{li2020federated}, federated learning can not only solve the problem of data islands but also has the advantages of protecting data privacy and alleviating communication pressure. At present, it has been combined with different fields and achieved satisfactory results, such as cyber-attack detection\cite{wu2020fedhome}, traffic flow prediction\cite{liu2020privacy}, spatiotemporal scenario generation of renewable energy\cite{li2022privacy}.  

\begin{figure}[t]
	\centering	
	\includegraphics[width=3.5in]{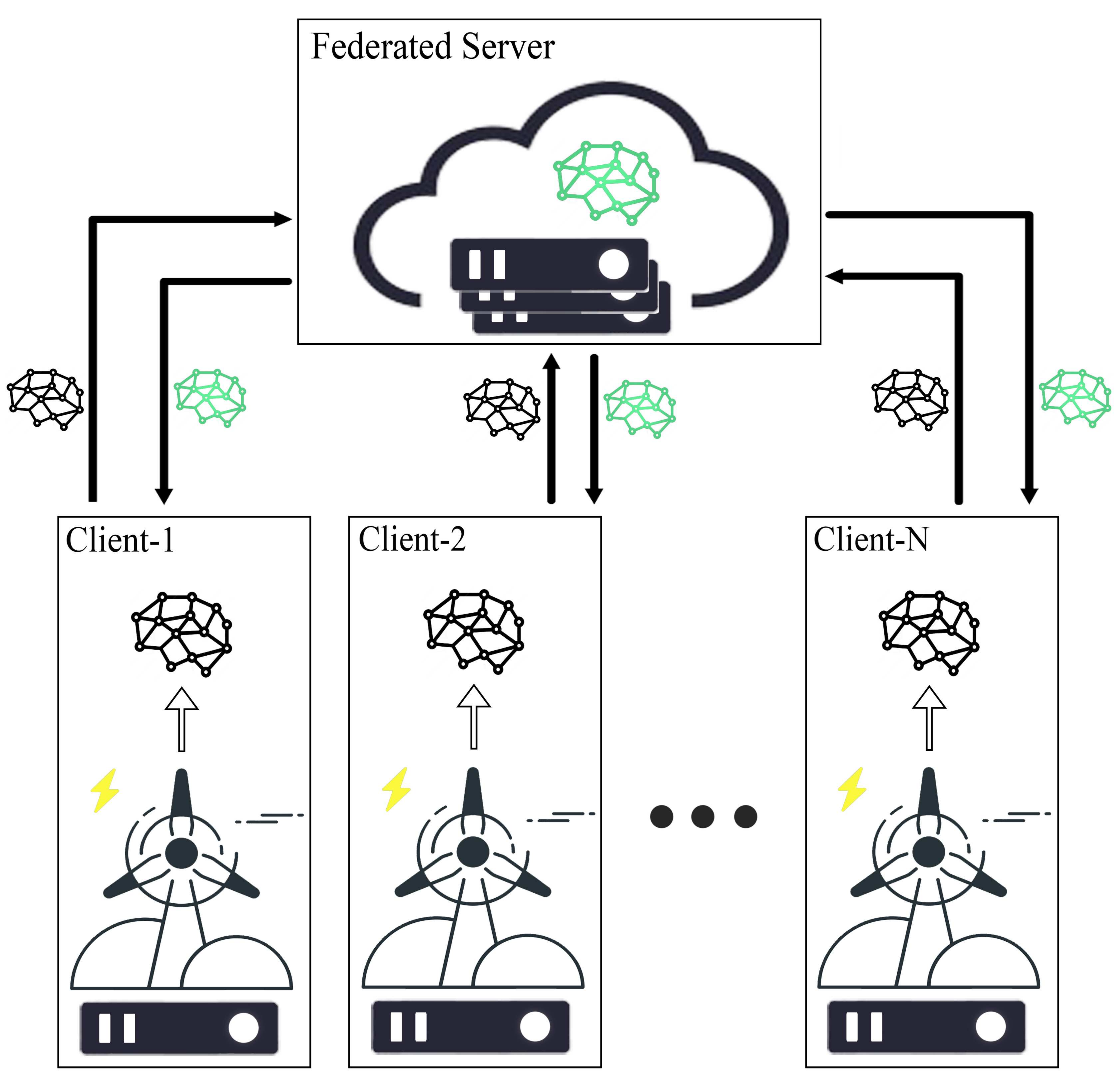}	
	\caption{{The basic structure of federated learning.}}	
	\label{fig.FL}
\end{figure}

Federated learning is mainly composed of a federated server and multiple clients. Unlike the traditional centralized method that requires the data of each client to be gathered in a centralized workstation, most of the training process in federated learning is done locally on each client's edge device. Therefore, the raw data is not shared between each other, it has a good privacy protection ability. Federated learning can solve the problems of data islands and privacy protection by sharing parameters, and its working principle is as follows: Firstly, an initial global model is allocated to each client by the federated server, and each client uses its private data to train the model locally. Secondly, when the local training is completed, the client uploads the weight parameters of each local model to the federated server, and then the parameters of the global model will be updated according to the collected parameters of each client. Finally, the federated server passes the updated global model parameters to each client again. Repeat the above process until the pre-set criteria are satisfied. The overall structure of federated learning is shown in Fig. \ref{fig.FL}.

The basic requirement of using the federated learning framework is to ensure the model's performance. Compared with the centralized training methods, the performance of the federated learning method needs to meet the following requirements.
\begin{equation}\label{eq.performance}
    \left| {{\psi _C} - {\psi _F}} \right| \textless \sigma
\end{equation}
where $\psi_C$ is the evaluation criteria  of the centralized model, $\psi _F$ is the evaluation criteria of the federated model, and $\sigma$ is a small non-negative number. It can be seen form (\ref{eq.performance}) that the performance of the model that uses the federated learning framework for distributed training is very similar to that of the model that aggregates all data for centralized training\cite{yang2019federated}.

\section{PROPOSED METHODOLOGY}\label{sec.PROPOSED METHODOLOGY}
The method of using DRL to solve the prediction problem is first explained in this section, and then the specific operation process of the proposed FedDRL is described in detail.
\subsection{Problem Formulation}
The highly fluctuating of wind power significantly affects the accuracy of wind power forecasting. To improve the forecasting accuracy and enhance the robustness of the forecasting model, we use the DDPG algorithm in DRL as the basic model for wind power forecasting in this work. 

To use DDPG for wind power prediction, this paper converts the prediction problem into a decision-making problem. We select the wind power for several consecutive time periods to construct a vector as follows, which is used as state, $s_t$.
\begin{equation}
    s_t = [P^w_{t-(j-1)}, P^w_{t-(j-2)}, P^w_{t-(j-3)}, P^w_{t-(j-4)}, ..., P^w_{j-1}, P^w_t]
\end{equation}
where $P^w$ is wind power, $j$ is the number of selected time periods. The agent makes a decision based on $s_t$, and outputs the predicted value of wind power as action, $a_t$, after which the agent will obtain the next state $s_{t+1}$ and reward $r$. For the wind power forecasting problem, the reward $r$ of the proposed FedDRL is defined as an error zone, not just the error between the model output and the label in traditional supervised learning. Therefore, the sensitivity of the prediction model to noise data can be reduced, and the robustness of the prediction model can be improved \cite{kuremoto2019training}. The reward function set in this work is as follows.
\begin{equation}\label{eq.reward function}
    r =  - \left| {{a_r} - {a_t}} \right|
\end{equation}
where $a_r$ is the true value of wind power. In the Actor part, the loss function is constructed to minimize the prediction errors. In the Critic part, the goal of the loss function is to continuously shrink the difference between the actual and the predicted reward.

After the above-mentioned method is used to transform the prediction problem into a decision-making problem, DDPG can be reasonably used as the wind power forecasting model in this work. The actor outputs the predicted value according to the observation, which is evaluated by the critic, and the neural networks are updated according to the temporal-difference error (TD-error), finally, a prediction model with good performance is obtained. The specific steps of using DDPG for wind power forecasting are as follows.

\emph{\textbf{Step 1:}} Construct a vector consisting of wind power for several consecutive time periods as the state. 

\emph{\textbf{Step 2:}} Predict a wind power value as an action based on the current state.

\emph{\textbf{Step 3:}} Calculate the difference between the predicted value and the real value, and take the negative number of its absolute value as a reward; then get the next state.

\emph{\textbf{Step 4:}} Update the actor and critic neural network, and a model with good prediction performance can be obtained through continuous training.

\subsection{FedDRL Framework}
Traditional centralized prediction methods may have data leakage problems in the process of data collection and storage, which involves sensitive privacy issues. The emergence of federated learning undoubtedly provides an effective solution to the above problems.

\begin{figure}[t]
	\centering	
	\includegraphics[width=3.6in]{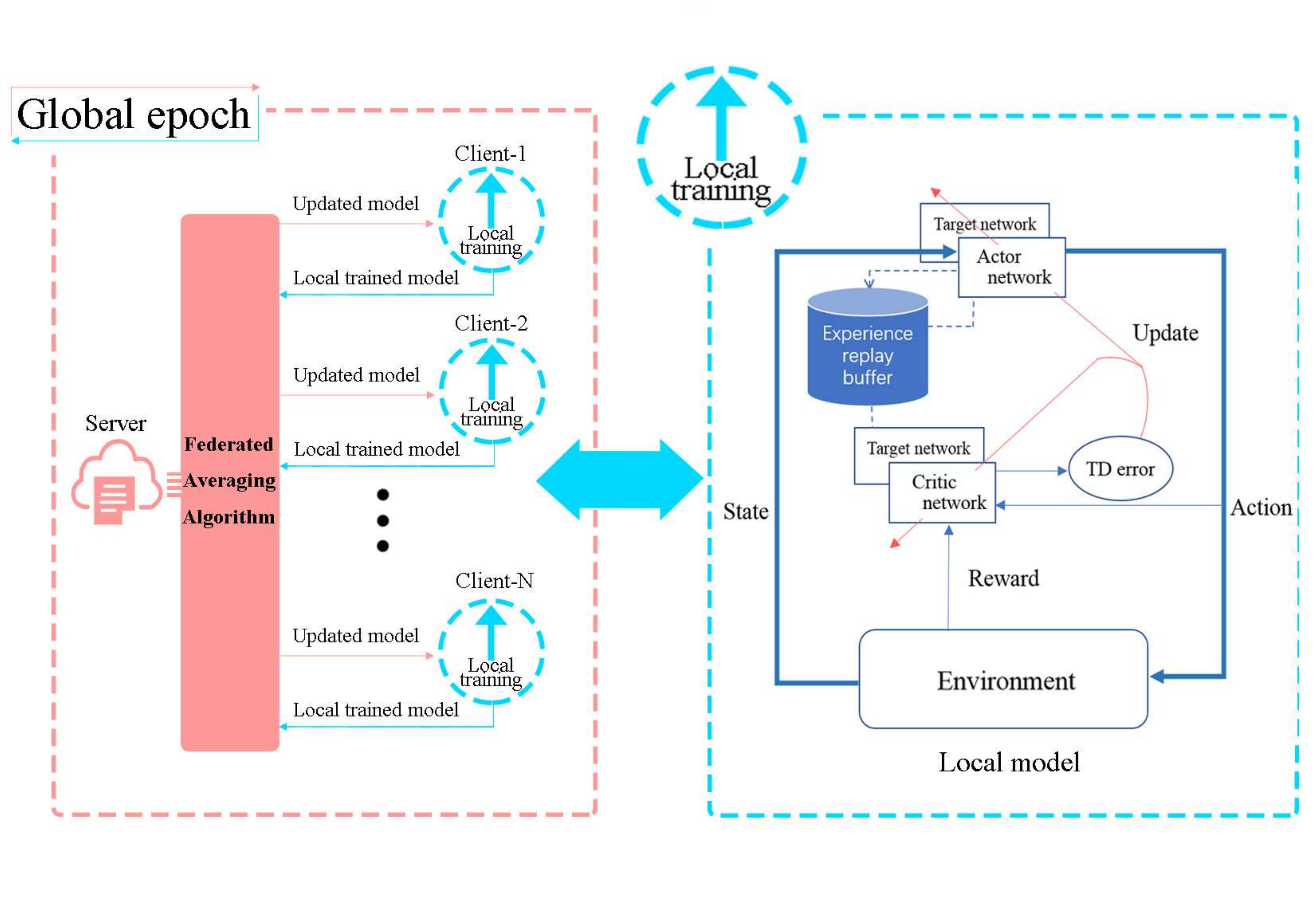}	
	\caption{{FedDRL framework.}}	
	\label{fig.FedDRL}
\end{figure}

We propose a FedDRL prediction scheme, which adopts federated learning as the overall framework, the federated server and each client hold a DDPG prediction model as the global model and the local model, respectively. The FedDRL forecasting scheme is mainly controlled by two parameters, the customer's participation ratio $E$ and synchronization interval $K$. We assume that each wind farm has the ability to independently train a global model. Considering that the computing power of each wind farm may be weak, the customer's local training only takes a small number of iterations. The framework of the proposed FedDRL is shown in Fig. \ref{fig.FedDRL}.

Since tuning the hyperparameters of the model is a very time-consuming and labor-intensive task, to simplify the deployment of FedDRL, we adopt an automatic machine learning method based on Bayesian optimization called metis to automatically select the hyperparameters of the global model \cite{li2018metis}. We set the actor and critic as a fixed-depth neural network, and let the metis tuner confirm the hyperparameters, such as hidden neurons and learning rate of the actor and critic network.

The federated averaging algorithm is the core mechanism for FedDRL to achieve user privacy protection and reduce communication pressure. The main idea of the federated averaging algorithm is to aggregate the model parameters trained locally by the participating clients in each round to the federated server, and then perform the averaging operation on them \cite{mcmahan2017communication}. In FedDRL, we collect the client parameters in each training round which is mainly about the trained parameters of the actor and critic networks in the local model and average them to update the global model. The process is expressed as

 \begin{equation}\label{eq.fedavg}
           \begin{array}{l}
\theta _S^Q \leftarrow \frac{1}{{{N_e}}}\sum\limits_{e = 1}^{{N_e}} {\theta _e^Q} ,{\rm{ }}\theta _S^\mu  \leftarrow \frac{1}{{{N_e}}}\sum\limits_{e = 1}^{{N_e}} {\theta _e^\mu } ,\\
\theta _S^{Q'} \leftarrow \frac{1}{{{N_e}}}\sum\limits_{e = 1}^{{N_e}} {\theta _e^{Q'},{\rm{ }}\theta _S^{\mu '} \leftarrow \frac{1}{{{N_e}}}\sum\limits_{e = 1}^{{N_e}} {\theta _e^{\mu '}} } 
\end{array}
        \end{equation}
where $N$ is the total number of clients, $e$ is the client number participating in the training epoch; $\theta _S^\mu$ and $\theta _S^Q$ are the main actor and critic network parameters of the global model, $\theta _S^{\mu '}$ and $\theta _S^{Q'}$ are the target actor and critic network parameters of the global model; $\theta _e^\mu$ and $\theta _e^Q$ are the main actor and critic network parameters of each local model, $\theta _e^{\mu '}$ and $\theta _e^{Q'}$ are the target actor and critic network parameters of each local model. After the global model is updated, the parameters will be returned to clients to update the local model.

\subsection{FedDRL-based Wind Power Forecasting}
We propose a prediction scheme that combines federated learning and DRL. On the basis of ensuring the performance of the prediction model, the proposed FedDRL can not only solve the problem of data islands among wind farms but also protect clients' data privacy. The specific process of FedDRL mainly includes the following four steps.

\emph{\textbf{Step 1:}} Initial model allocation. The local model in each wind farm and the global model are all initialized. The federated server obtains a global model based on public data with hyperparameters automatically selected by the metis tuner, and then pass the global model to participating wind farms.

\emph{\textbf{Step 2:}} Local model training and parameters upload. Each wind farm updates the local model according to the received global model and then uses their private data to train the model locally. After the training is completed, each wind farm uploads the trained weight information of the actor and critic network to the federated server.

\emph{\textbf{Step 3:}} Global model update. The federated server gathers the weight information uploaded by each wind farm and uses the FederatedAveraging algorithm to update the global model. The updated global model is re-distributed to each client.

\emph{\textbf{Step 4:}} Wind power forecasting. Repeat \emph{\textbf{Steps 2}} and \emph{\textbf{3}} until the global model meets the required criteria and stops training. The federated server transmits the trained model to each wind farm, and each wind farm uses it to forecast wind power.

\begin{algorithm}[t] 
\small
\color{black}
\renewcommand{\algorithmicrequire}{\textbf{Require:}}
\renewcommand{\algorithmicensure}{\textbf{Require:}}
    \caption{Federated Deep Reinforcement Learning (Fed-DRL)} 
    \label{alg:fed}
	\begin{algorithmic}[1] 
    \REQUIRE Local episode $M$; Global epoch $W$; Synchronization interval $K$, Client ratio $E$.
    \ENSURE A global DDPG model with parameters ($\theta _S^Q,\theta _S^\mu$ ) , ($\theta _S^{Q'},\theta _S^{\mu '}$) for main critic network and actor network , target critic network and actor network on central server $S$; local DDPG model with parameters $\{ \theta _{}^Q,\theta _{}^\mu \} _{i = 1}^N$, $\{ \theta _{}^{Q'},\theta _{}^{\mu '}\} _{i = 1}^N$ for main critic network and actor network, target critic network and actor network on $N$ client  $\{ C\} _{i = 1}^N$
    \FOR{global epoch $w=1,2,\ldots, W$}
    \STATE Randomly select $N_e$ clients from all clients $ \{ C\}$ using $E$
    \FOR{each selected client $e$ \textbf{in parallel}} 
    \color{black}
    \FOR{episode $m= 1,2,\ldots,M$ }
    \STATE Initialize a random process $H$ for action exploration.
    \STATE  Receive initial observation state $s$.
    \FOR{$t=1,2,\ldots, T$}
    \STATE Select action $a$ according to the current policy and exploration noise.
    \STATE Execute action $a$ and get reward $r$ and new state $s'$.
    \STATE Store transition ($s, a, r, s'$) in replay buffer.
    \STATE Sample a minibatch from replay buffer.
    \STATE Update main actor network by (\ref{eq.update main actor network})
    \STATE Update main critic network by (\ref{eq.update the main critic network})
    \STATE Update the target networks by (\ref{eq.update target networks})
    \ENDFOR
    \ENDFOR
    \ENDFOR
    \IF {$w \mod K = 0$} 
        \STATE All selected clients send parameters to server, and the server update the parameters of the global model by (\ref{eq.fedavg}).
        \STATE The server send back parameters and clients update local parameters
        \begin{equation*}
           \{ \theta _{}^Q,\theta _{}^\mu \} _{i = 1}^N \leftarrow (\theta _S^Q,\theta _S^\mu ),{\rm{ }}\{ \theta _{}^{Q'},\theta _{}^{\mu '}\} _{i = 1}^N \leftarrow (\theta _S^{Q'},\theta _S^{\mu '})
        \end{equation*}
    \ENDIF
    \ENDFOR
  \end{algorithmic}
\end{algorithm}

The complete FedDRL algorithm is shown in Algorithm \ref{alg:fed}.

\begin{figure*}[t]
	\centering	
	\includegraphics[width=7.2in]{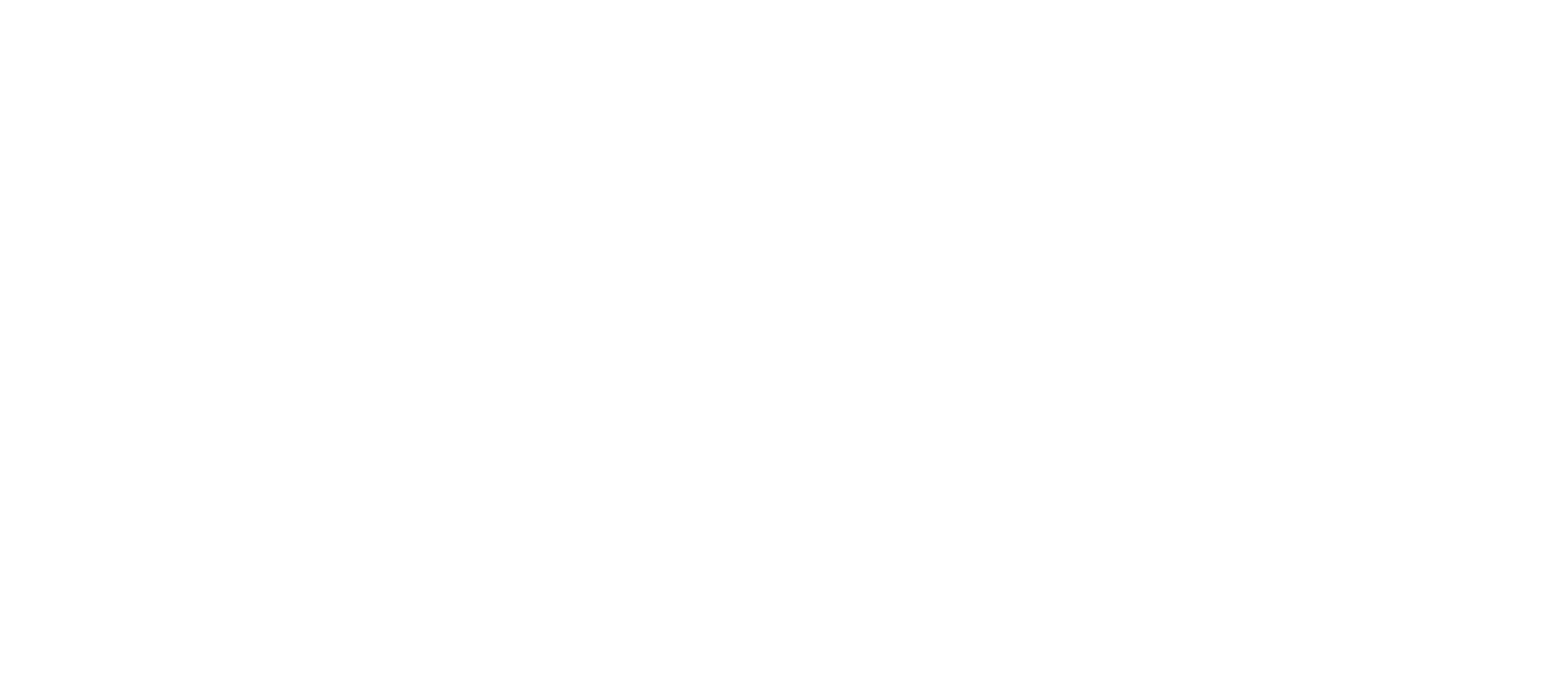}	\caption{{\color{black}Forecasting results of different wind farms.}}	
	\label{fig_4results}
\end{figure*}

\subsection{Evaluation Indices}
This paper uses the more common normalized mean absolute error (NMAE) and normalized root mean square error (NRMSE) as evaluation indicators to assess the forecasting performance of the proposed forecasting scheme\cite{zhao2016novel}.

\begin{equation}
    \begin{aligned}
     NMAE = \frac{1}{V}[\sum\nolimits_{i = 1}^V {(|y_i^{n} - {p_i^{n}}|)]} {\rm{ }}
    \end{aligned}
\end{equation}

\begin{equation}
    \begin{aligned}
      NRMSE = \sqrt {\frac{1}{V}\sum\limits_{i = 1}^V {{{\left( {y_i^{n} - p_i^{n}} \right)}^2}} } 
    \end{aligned}
\end{equation}

\begin{equation}
    y_i^n = \frac{{{y_i}}}{{\max _{i = 1}^V{y_i}}}
\end{equation}

\begin{equation}
    p_i^n = \frac{{{p_i}}}{{\max _{i = 1}^V{y_i}}}
\end{equation}
where $V$ denotes the total number of samples, $i$ represent the samples number; $y$ denotes the real value, $p$ is the prediction result; $y^n$ and $p^n$ are the normalized value of $y$ and $p$.

\subsection{Networking Load Gain}\label{sec.load gain}
The proposed FedDRL forecasting scheme can not only protect the data privacy of each client but also has the advantage of relieving communication pressure. To verify the communication advantages of the FedDRL, we define the network load of the centralized forecasting method and the proposed forecasting scheme as follows,  respectively.
\begin{equation}
    {L_C} = \sum\limits_{n = 1}^N D  \times {h_n}\ , \ \ \ {L_F} = I \times \sum\limits_{w = 1}^W {} \sum\limits_{n = 1}^N {{h_n}}
\end{equation}
where $D$ is the size of the private data of each client in the centralized method, $I$ is the size of the model information that each client needs to transmit in FedDRL, and $h$ is the hops between each client and the central server. Then the networking load gain $U$ is defined as follows.
\begin{equation}
    U = 1 - {{{L_F}} \mathord{\left/
 {\vphantom {{{L_F}} {{L_C}}}} \right.
 \kern-\nulldelimiterspace} {{L_C}}}
\end{equation}

\begin{figure}[t!]
	\centering	
	\includegraphics[width=3.6in]{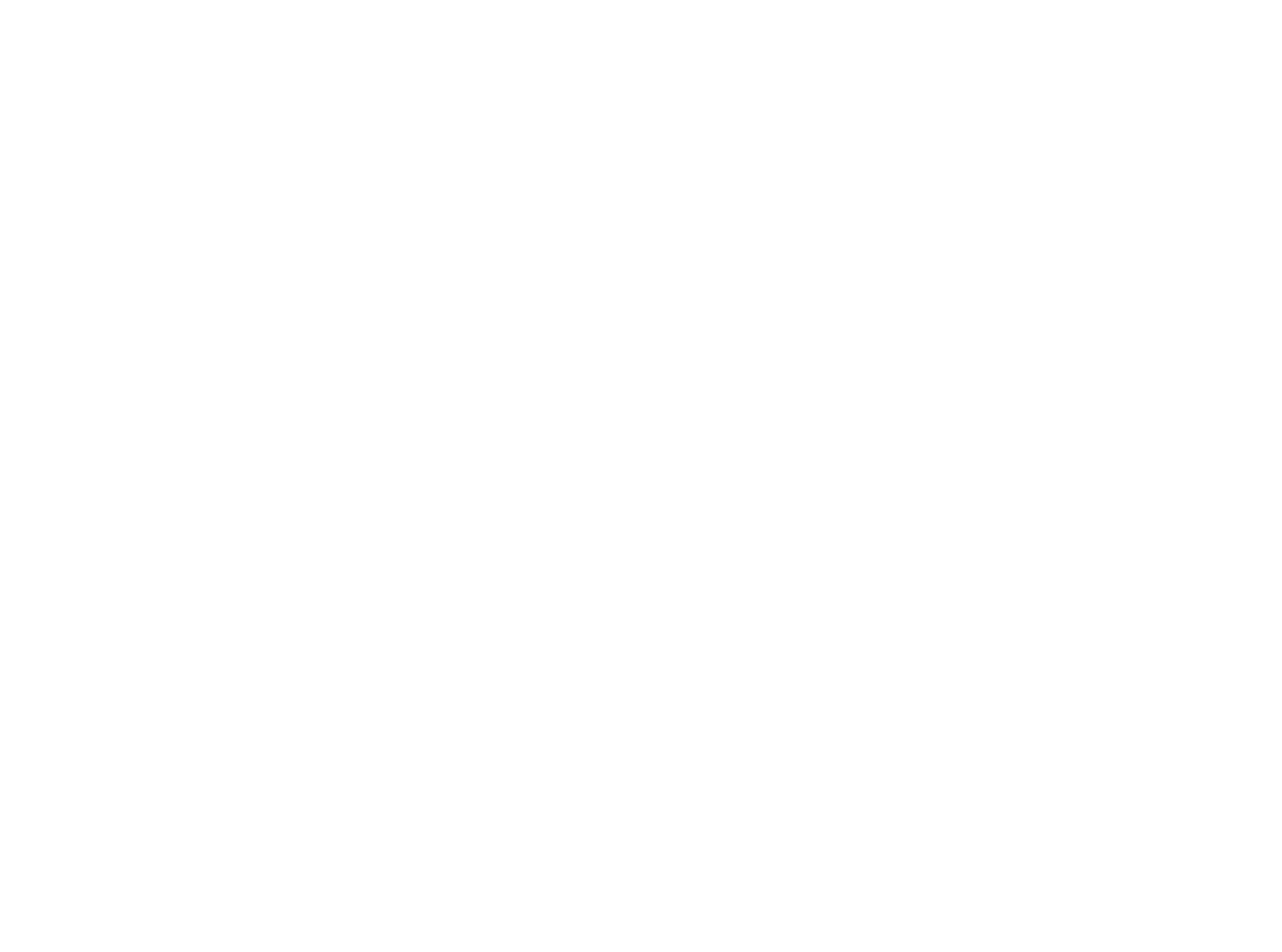}	\caption{{\color{black}Probability distribution of forecasting errors}}	
	\label{fig_pdf}
\end{figure}

\begin{figure}[t]
	\centering	
	\includegraphics[width=3.6in]{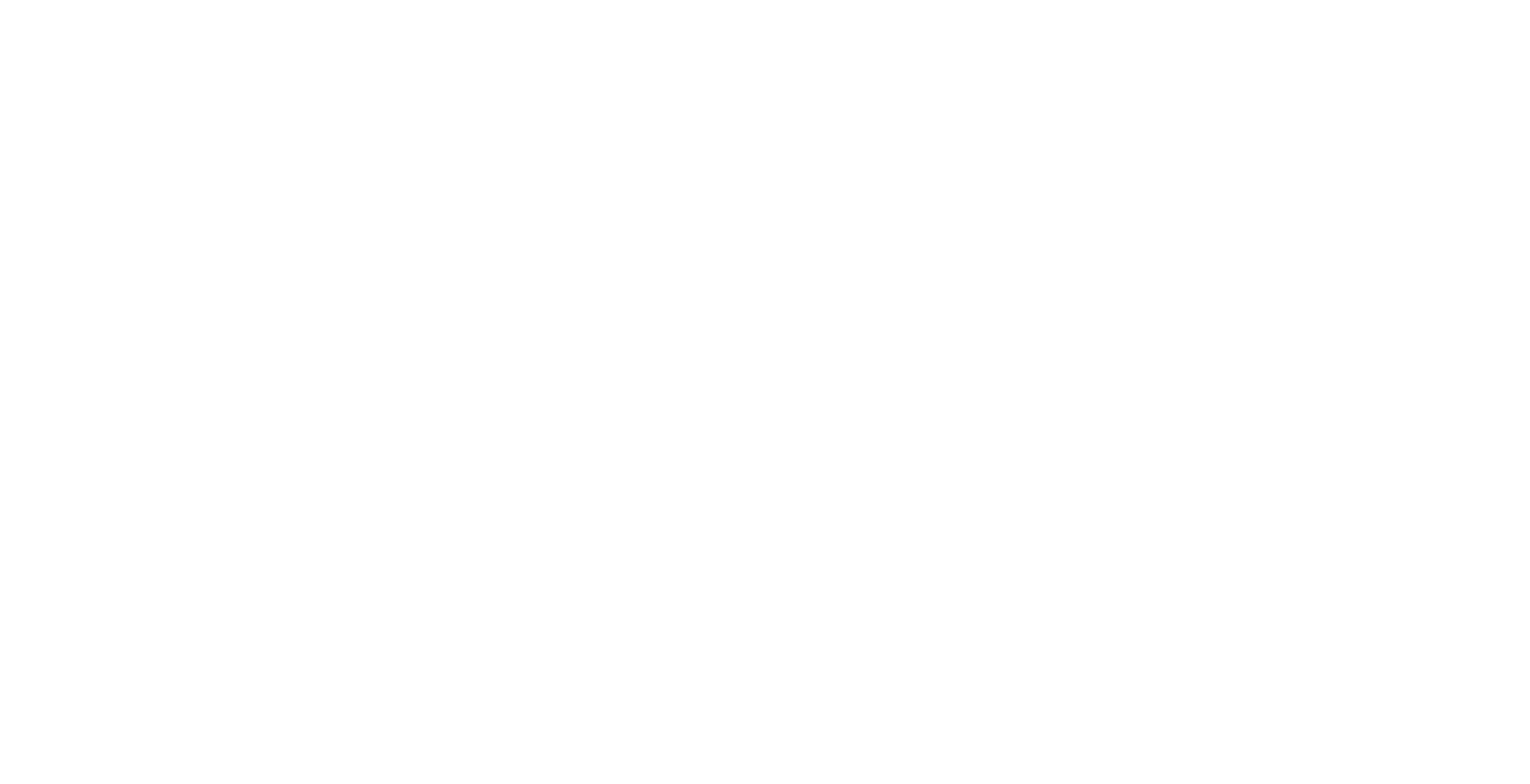}	\caption{{\color{black}Changes in reward value during training}}	
	\label{fig_prossess}
\end{figure}

\begin{figure}[t]
	\centering	
	\includegraphics[width=3.6in]{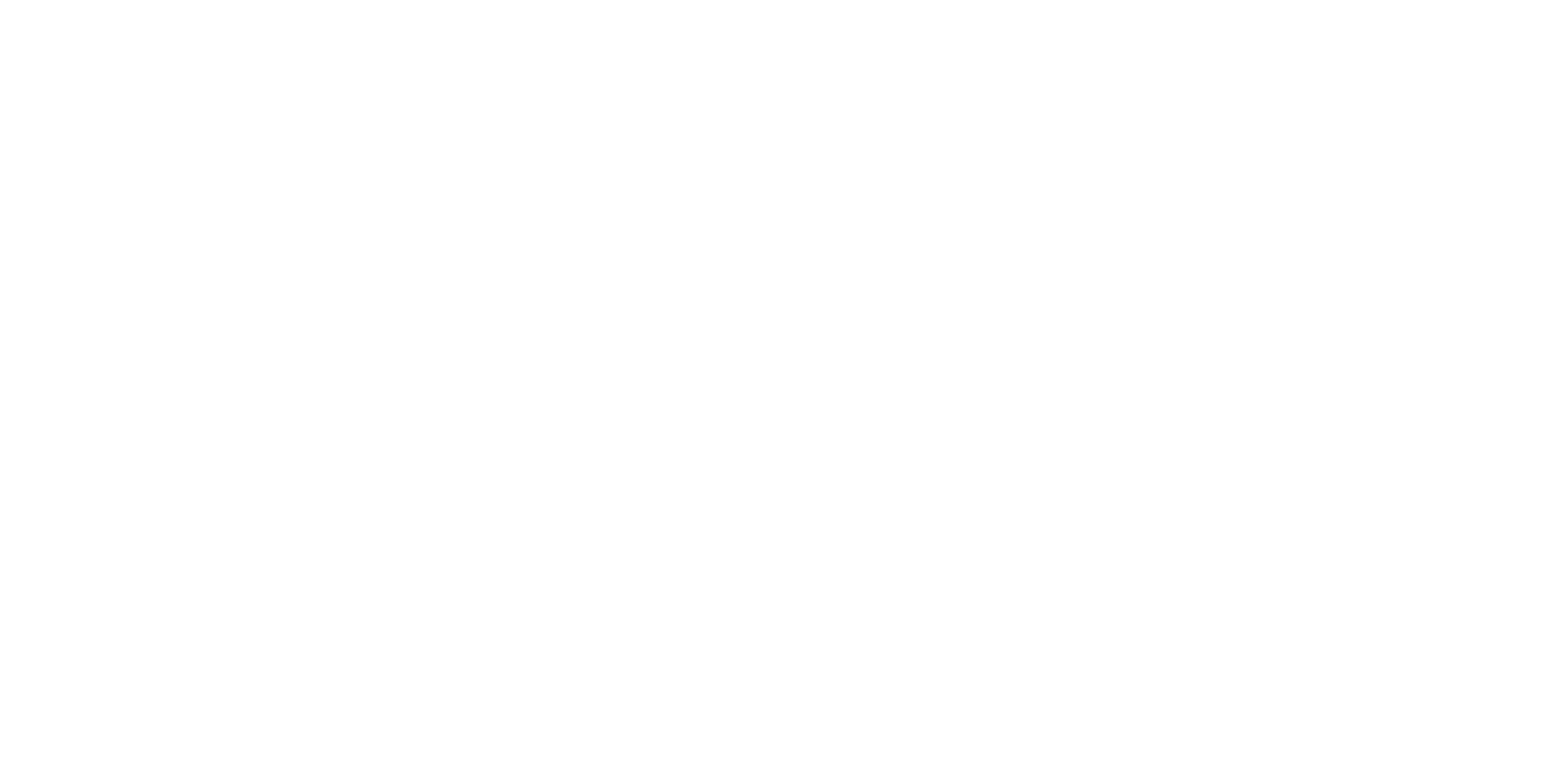}	\caption{{\color{black}The simulations results with different federated parameters setting}}	
	\label{fig_KE}
\end{figure}

\section{CASE STUDY}\label{sec.CASE STUDY}
This article uses real-world historical data for simulations and verifies the effectiveness and superiority of the proposed forecasting scheme by comparing the predicted results with real data. In addition, we also designed experiments on the influence of different federal parameter settings on FedDRL to verify the robustness of the proposed forecasting scheme. In this work, all the simulations are implemented with Python 3.7, Tensorflow 2.0 on a PC platform with 2 Intel Core dual-core CPUs (2.6Hz) and
6 GB RAM.

\subsection{Forecasting Model Parameter Settings}
To improve the efficiency of model construction, this paper applies automatic machine learning to the proposed scheme based on our team’s previous research\cite{li2021optimal1}. The appropriate hyperparameters for the forecasting model are chosen automatically. In this study, $j$ is set to 7 after conducting various simulations, since it can achieve satisfactory prediction performance in most cases. Federated parameters $K$ and $E$ are set to 100 and 100\%, respectively. The main model parameters are shown in Tab. \ref{tab.parameters of feddrl}.

\begin{table}[t]
  \centering
  \caption{Hyperparameters of FedDRL}\label{tab.parameters of feddrl}
    \begin{tabular}{cccc}
    \toprule
          & Learning rate & Hidden neurons & Activation function \\
    \midrule
    Actor & 0.0003 & 30    & Relu/Sigmoid \\
    Critic & 0.003 & 28    & Relu/Linear \\
    \bottomrule
    \end{tabular}%
  \label{tab:addlabel}%
\end{table}%

\subsection{Data Description} 
In this work, we use the real-world datasets consisting of 11 wind farms with a sampling interval of 5 minutes in Washington State, USA to be the local datasets of clients, which are gained from National Renewable Energy Laboratory (NREL). The forecasting horizon is 5 minutes. When each wind farm uses a separate private dataset for local training, the ratio of training set to testing set is 4:1, which is in accordance with the setting in \cite{li2022privacy}.

\begin{figure*}[t]
	\centering	
	\includegraphics[width=5.8in]{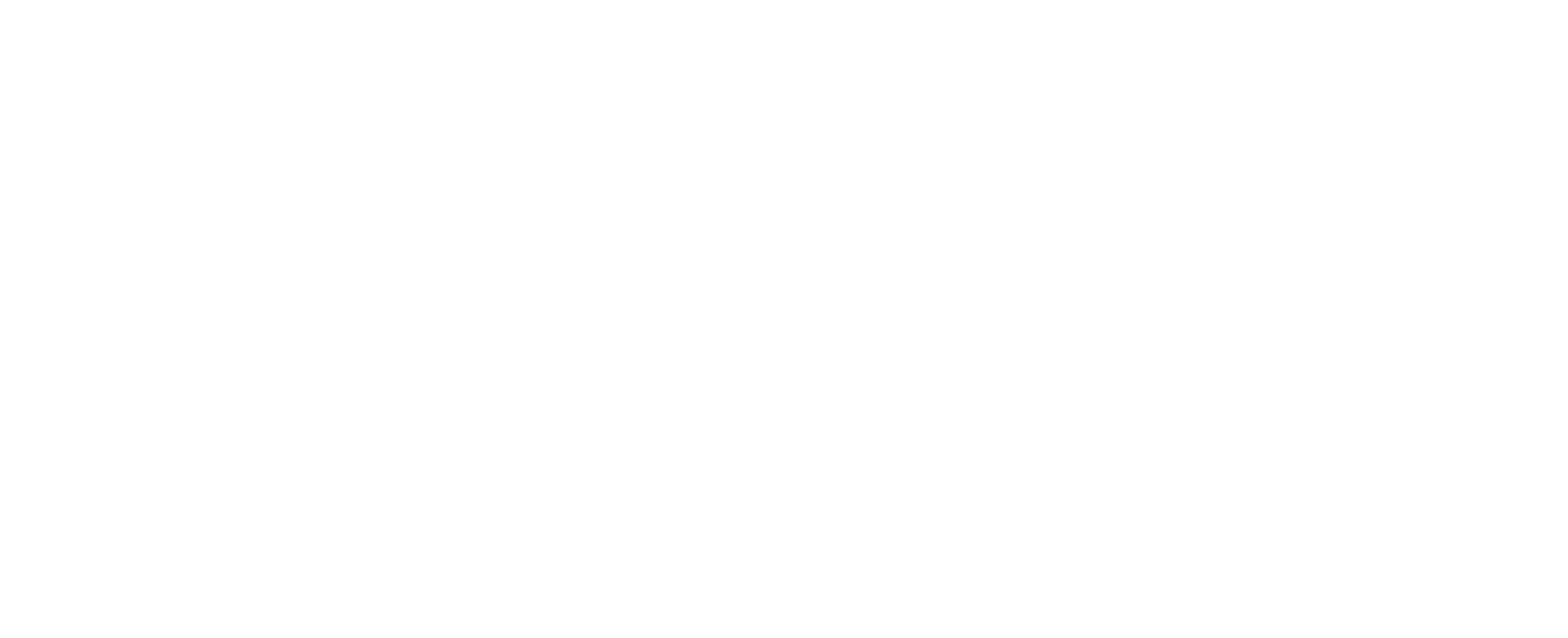}	\caption{{\color{black}Forecasting results of the two methods}}	
	\label{fig_FvsC}
\end{figure*}

\subsection{Forecasting Results and Performance Analysis}
\subsubsection{Forecasting results of different wind farms}
Fig. \ref{fig_4results} shows the forecasting results of four wind farms participating in the proposed FedDRL forecasting scheme. It can be seen that the final prediction results of each wind farm can exactly describe the real wind power output, which shows that the use of the proposed FedDRL scheme can eliminate the problem of data islands and enable different stakeholders to gain an accurate forecasting model without sharing their private data.

\subsubsection{Forecasting errors probability distribution of different wind farms}
The probability density diagram of the forecast errors of four wind farms is demonstrated in Fig. \ref{fig_pdf}. The total area enclosed by the curve and the coordinate axis in each figure is 1, which represents the range of higher and lower probability of forecasting errors. From the figure, we can find that the prediction errors of each wind farm are concentrated in a small range, which certifies that each client can achieve good prediction results relatively stably, and it also proves that the DDPG local prediction model has good robustness.

\subsubsection{Forecasting performance comparison between wind farms}
All clients participating in the proposed FedDRL scheme hope to achieve accurate predictions without sharing private data because of the concerns of privacy protection. Tab. \ref{tab.Prediction performance comparison} shows the evaluation indices of 4 randomly selected clients. We can learn from the table that although the forecasting errors of each wind field are different, the overall prediction results are relatively accurate, indicating that the proposed scheme can provide a good performance prediction model to the participating clients and protect the clients' privacy.

\begin{table}[t]
  \centering
  \caption{Prediction performance comparison}\label{tab.Prediction performance comparison}
    \begin{tabular}{crrrr}
    \toprule
    \textcolor[rgb]{ .067,  .067,  .067}{Indicators} & \multicolumn{1}{c}{\textcolor[rgb]{ .067,  .067,  .067}{Wind farm 1}} & \multicolumn{1}{c}{\textcolor[rgb]{ .067,  .067,  .067}{Wind farm 2}} & \multicolumn{1}{c}{\textcolor[rgb]{ .067,  .067,  .067}{Wind farm 3}} & \multicolumn{1}{c}{\textcolor[rgb]{ .067,  .067,  .067}{Wind farm 4}} \\
    \midrule
    \textcolor[rgb]{ .067,  .067,  .067}{NMAE} & 0.020651404
 & 0.021026766
 & 0.039084657 & 0.025438397
 \\
    \textcolor[rgb]{ .067,  .067,  .067}{NRMSE} & 0.037752126
 & 0.029915878
 & 0.055905707
 & 0.047609882
 \\
    \bottomrule
    \end{tabular}%
  \label{tab:addlabel}%
\end{table}%

\subsubsection{Training process analysis of the wind farms}
Fig. \ref{fig_prossess} depicts the change process of reward values of four randomly selected clients during training in our proposed FedDRL. The reward value of all clients fluctuates sharply in the initial stage of the iterative process, which means that each local model is constantly exploring various policies that can achieve good prediction effects when updating model parameters and performing local training; after a certain number of global epoch and local training, the prediction models of each client begin to converge, and the reward value tends to stabilize, indicating that the forecasting model has gradually explored a suitable policy and can accurately predict the wind power.

\subsubsection{Robustness test of FedDRL}
We conduct a simulation by changing the synchronization intervals $K$ and the clients' participating ratio $E$ to examine the robustness of the proposal.

Fig. \ref{fig_KE} exhibits the influence of different combinations between synchronization intervals of 50, 100, 200, and participation rates of 50\% and 100\% on the FedDRL training process. It can be learned from the figure that with different combinations of $K$ and $E$, the error of the proposed forecasting scheme will eventually converge to a small range. More importantly, when the participation ratio of clients is fixed, synchronization intervals have barely effect on the prediction process and results of the scheme; while for a given synchronization intervals, the participation ratio of clients will only slightly affect the convergence speed, and has no effect on the final prediction performance of the scheme. It can be concluded that the proposed forecasting scheme has good robustness to different federated parameters variations.

\begin{table}[t]
  \centering
  \caption{Forecasting performance of different methods}\label{tab.performance of different methods}
    \begin{tabular}{ccc}
    \toprule
    Method & NMAE  & NRMSE \\
    \midrule
    Proposed method & 0.0206514 & 0.0377521 \\
    RDPG  & 0.0316357 & 0.08384312 \\
    ARIMA & 0.0291881 & 0.07052312 \\
    BPNN  & 0.0790775 & 0.20001125 \\
    MLR   & 0.0924137 & 0.23703316 \\
    \bottomrule
    \end{tabular}%
  \label{tab:addlabel}%
\end{table}%

\subsection{Comparative Analysis of FedDRL and Traditional Centralized Forecasting Methods}

\subsubsection{Performance comparison of different forecasting methods}
This paper compares the prediction performance of the proposed prediction method with classic ARIMA, mixed logistic regression (MLR) prediction model, typical back propagation neural network (BPNN), and another reinforcement learning algorithm recurrent deterministic policy gradient (RDPG). We use FedDRL and other forecasting methods to predict the wind power of the same wind farm, and the centralized forecasting method is used by the comparison prediction approaches. In the ARIMA, the parameters of lag observations $p$, the number of times that the raw observations are differenced $d$, and the size of the moving average window $q$ are set to 2, 0, and 1. The hidden neurons and layers are set to 10 and 2 in the BPNN, respectively. The hidden neurons of Actor and Critic network in the RDPG are set to 34 and 36.

The results shown in Tab. \ref{tab.performance of different methods} suggest that judging from the performance indicators of various prediction methods, the error of the proposed method is smaller than that of other methods and the NMAE and NRMSE can reduce at least 29.25\% and 46.47\%, respectively, which proves that the  FedDRL does have the advantage in prediction accuracy.

\subsubsection{Forecasting results of FedDRL and centralized DDPG}

The wind power prediction results of the same wind farm using FedDRL and centralized DDPG models are shown in Fig. \ref{fig_FvsC}. Fig. \ref{fig_FvsC}(a) is the comparison between the wind power predicted by the two methods and the true value, Fig. \ref{fig_FvsC}(b) is the probability density of the forecasting wind power. 

The figure demonstrates that both the proposed FedDRL and the centralized DDPG show good forecasting performance, which can accurately describe the wind power output of the wind farm. The accuracy of wind power prediction using the proposed FedDRL is comparable to that using centralized DDPG. It is worth noting that the use of centralized methods for forecasting requires gathering data from various wind farms, which may involve sensitive user privacy issues. When the private data of multiple wind farms are gathered together, once the privacy data is leaked, it may cause very serious losses to each wind farm. While the method proposed in this paper can save the data of each wind farm locally for training, and achieve a similar prediction accuracy without transmitting raw data, which suggests that the proposed FedDRL can achieve the forecasting accuracy criterion without sacrificing user privacy.

\subsubsection{Networking load gain}
For ease of simulation, in this work, we assume that the distance between all clients and the server is 1-Hop. According to the formula in sec.\ref{sec.load gain}, we can calculate that when the synchronization interval is 50, 100, 200, the network load gain using the designed FedDRL is at least 88.34\%, 94.82\%, 97.08\% compared with the traditional centralized prediction method. This verifies that the proposed prediction scheme can significantly relieve the communication pressure. When the proposed method covers a wider area and has more participating clients, a better networking load gain will be obtained.

\section{conclusion}\label{sec.conclusion}
As a clean and renewable energy source, wind power has developed rapidly in recent years. The development of advanced and accurate forecasting methods can undoubtedly promote the large-scale grid connection of wind power and assist the operation of power system. To deal with data privacy and openness, this paper proposes a FedDRL scheme for ultra-short-term wind power forecasting which combines federated learning and DRL. The DDPG algorithm is used as the basic prediction model of the FedDRL scheme to improve the forecasting accuracy; compared with the traditional wind power forecasting method, the framework of federated learning enable each stakeholder in the proposed forecasting scheme to upload only relevant parameters and save the private data locally for training. The proposed scheme not only guarantees the accuracy of wind power forecasting, solves the problem of data islands, but also effectively protects user privacy and relieves communication pressure.

The results of the simulations demonstrate that the designed FedDRL scheme outperforms the traditional wind power prediction methods with stable and more accurate forecasting results in a privacy-preserving way. The NMAE of most wind farms is in a very small range from 0.0207 to 0.0433. Compared with the traditional centralized prediction method, the proposed prediction scheme can obtain at least 88.34\% network load gain. In addition, the experiments of setting different participation rates and synchronization intervals in the federation parameters also proved that the proposed scheme has good robustness.

In future work, we will focus on how to improve the training efficiency of the model and apply the proposed scheme to the real-time dispatch of power system with renewables. Furthermore, this work does not consider the effects of packet loss rate and communication delay, how to solve these problems to facilitate the industrial application of the proposed method is also an interesting topic.

\bibliography{./References/ref}

\bibliographystyle{IEEEtran}

\end{document}